\pdfoutput=1
\documentclass[runningheads]{llncs}
\usepackage{algorithm}
\usepackage[noend]{algpseudocode}
\usepackage{stackengine}
\setstackEOL{\\}
\usepackage{bm}
\usepackage{graphicx}
\usepackage{float}
\usepackage{caption}
\usepackage{subcaption}
\usepackage{amsmath}
\usepackage{amssymb}
\usepackage [english]{babel}
\usepackage{spalign}
\usepackage [autostyle, english = american]{csquotes}
\usepackage{xcolor}
\usepackage[backend=bibtex]{biblatex}
\MakeOuterQuote{"}
\newcommand{\comment}[1]{}

\begin{document}
\title{Spirometry-based airways disease simulation and recognition using Machine Learning approaches}
\titlerunning{Tidal breathing disease simulation and recognition using ML}
%
\author{Riccardo Di Dio\inst{1, 2} \and
André Galligo \inst{1, 2} \and
Angelos Mantzaflaris\inst{1, 2} \and
Benjamin Mauroy\inst{2}}
\authorrunning{Riccardo Di Dio et al.}
%
\institute{${}^1$Université Côte d’Azur, Inria, France\\
${}^2$Université Côte d’Azur, CNRS, LJAD, VADER Center, France}
\maketitle              
\begin{abstract}
The purpose of this study is to provide means to physicians for automated and fast recognition of airways diseases. In this work, we mainly focus on measures that can be easily recorded using a spirometer. The signals used in this framework are simulated using the linear bi-compartment model of the lungs. This allows us to simulate ventilation under the hypothesis of ventilation at rest (tidal breathing). By changing the resistive and elastic parameters, data samples are realized simulating healthy, fibrosis and asthma breathing. On this synthetic data, different machine learning models are tested and their performance is assessed. All but the Naive bias classifier show accuracy of at least 99\%. This represents a proof of concept that Machine Learning can accurately differentiate diseases based on manufactured spirometry data. This paves the way for further developments on the topic, notably testing the model on real data.


\end{abstract}
\section{Introduction}
Having a fast and reliable diagnosis is a key step for starting the right treatment on time; towards this goal, Machine Learning (ML) techniques constitute potential tools for providing more information to physicians in multiple areas of medicine. More specifically, as far as respiratory medicine is concerned, there is a recent blooming of publications regarding the investigations of Artificial Intelligence (AI), yet the majority of them refers to computer vision on thoracic X-Rays or MRI \cite{Gonem2020}.
However, for lung diseases using Pulmonary Function Tests (PFTs) recent studies have only scratched the surface of their full potential, by coupling spirometry data with CT scans for investigating Chronic Obstructive Polmunary Diseases COPD on large datasets like COPDGene~\cite{Bodduluri20}. In our study, only normal ventilation is used allowing diagnosis also for children. Our aim is to provide a first proof-of-concept and provide the first positive results that could lead to fast, accurate and automated diagnosis of these diseases, similarly e.g. to Cystic Fibrosis (CF) where Sweat chloride test is a central asset \cite{Farrell2017}.

Normally, ML models are trained and tested on data and labels are provided by medical doctors. However, the originality of this study consists in using mathematical equations to simulate the ventilation following the directives of IEEE \cite{Hao2019}, then this data will be used to train the ML models. The obtained volume flows respect the expectations for both healthy and not healthy subjects. Using a synthetic model with a low number of parameters allows us to have everything under control. 

During this study, the lungs are modeled as elastic balloons sealed in the chest wall and the airways are modeled as rigid pipes, this allows to play with few parameters to simulate healthy and not healthy subjects and create synthetic volumetric data of tidal breathing. This data is then split and used to train and test different ML models for diagnosis. The accuracies reached during the study are very high, however, the choice of the parameters and the restriction of synthetic data allowed for promising results. Further tests are needed with real data to validate the accuracy reached. Nevertheless, this study points out that not every classifier is suited for this task.

A brief introduction to human lungs and its physiology is given in \ref{section:Ventilation}, then section \ref{section:MathModeling} shows how ventilation has been modeled. In section \ref{section:DataSetCreation} is shown how the dataset has been realized, section \ref{subsection:Training} reflects the training of the models on the dataset and finally in section \ref{section:Results} and \ref{section:Discussion} the results are exposed and discussed.
\subsection{Lung ventilation} \label{section:Ventilation}
The respiratory system can be split in two different areas, the bronchial tree (also referred as \textit{conducting zone}) and the acini, the \textit{respiratory zone}. The very beginning of the bronchial tree is composed of the trachea which is directly connected to the larynx, the mouth and the nose. The trachea can be seen as the \textit{root} of our tree, which then ramifies into two bronchi that will split again and again until about 23 divisions \cite{WeibelThePathwayForOxygen}. During each division, the dimensions of the children are smaller compared to the parent, according to Weibel's model, the reduction factor between each split is around 0.79 \cite{WeibelBook,Horsfield68,Tawhai2004,Mauroy2004}. After the very first ramification, the two bronchi leads to the left and the right lung. Inside the lungs, the ventilation takes place. The lungs are inflated and deflated thanks to the respiratory muscles. Their role is to transport the air deep enough in the lung so that the gas exchanges between air and blood could occur efficiently.

Some diseases affect the physiological behavior of the bronchial tree and of the lungs. In asthma, a general shrinking of the bronchi happens and the patient feels a lack of breath due to the increased total resistance of the bronchial tree. In mechanical terms, the patient will need a greater muscular effort in order to provide adequate pressure for restoring a normal flow within the lungs. 

In cystic fibrosis, there is an accumulation of mucus within the bronchial tree that will impact the capacity of the lungs to inflate and deflate, hence its rigidity. Normally, it is harder to breath for patients with cystic fibrosis because of the increased rigidity of their lungs.

The lungs mechanical properties can be used to build a mathematical model that can mimic the respiratory system.

\subsection{Mathematical modeling} \label{section:MathModeling}
\begin{figure}
\centering
\begin{subfigure}{.5\textwidth}
  \centering
  \includegraphics[width=.9\linewidth]{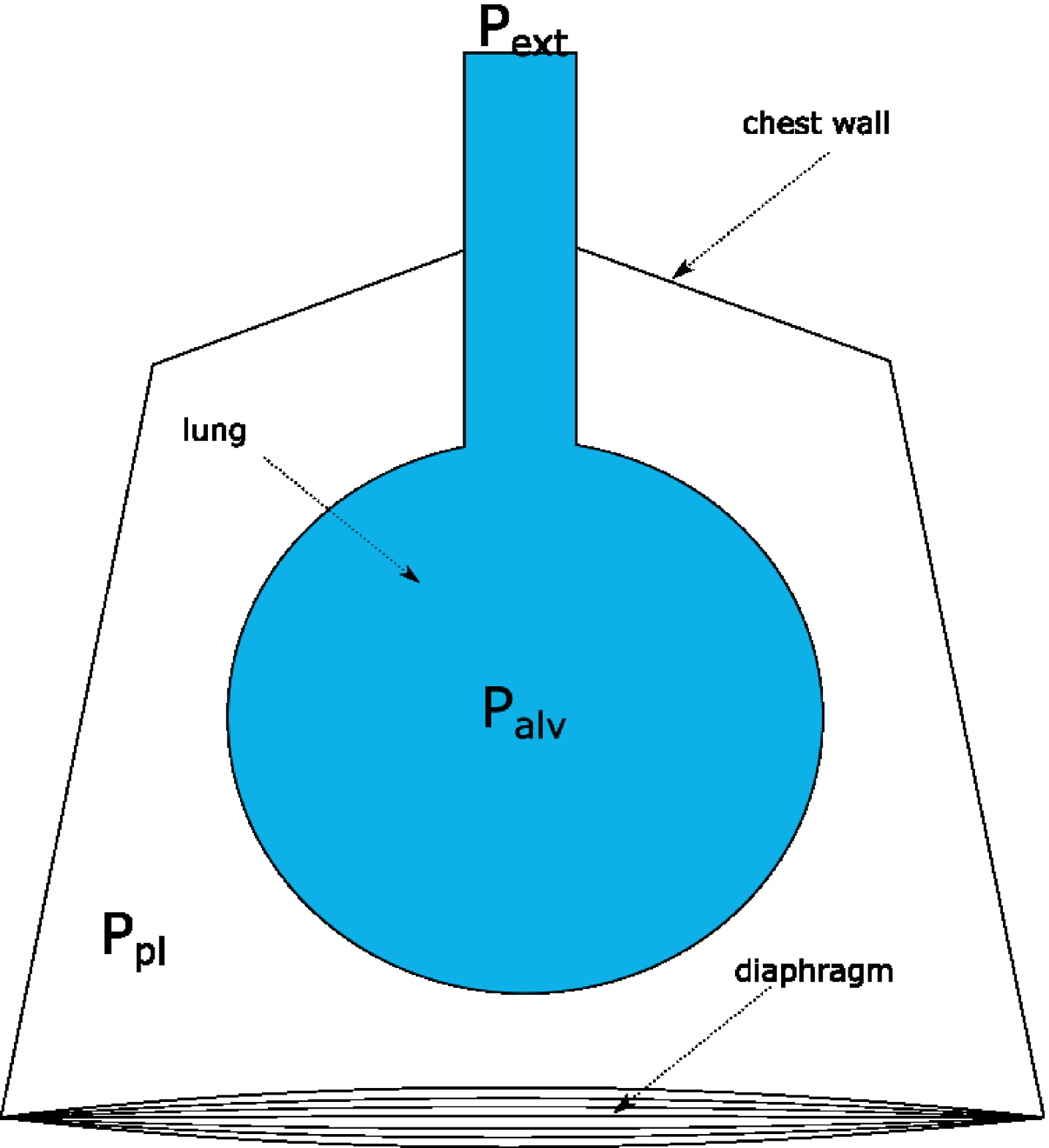}
  \caption{Mono-compartment model}
  \label{fig:MonoCompartment}
\end{subfigure}%
\begin{subfigure}{.5\textwidth}
  \centering
  \includegraphics[width=.9\linewidth]{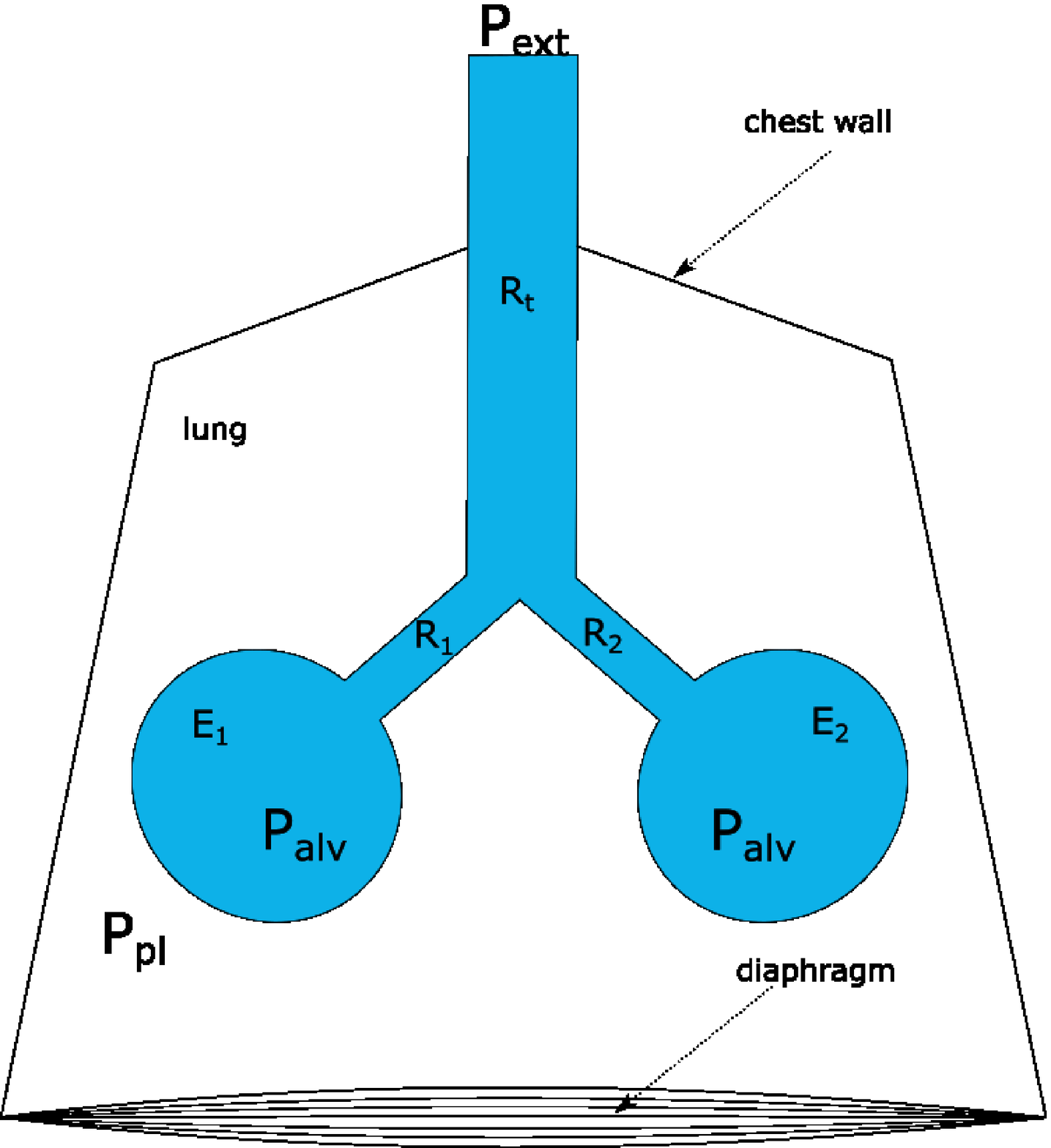}
  \caption{Bi-compartment model}
  \label{fig:BiCompartment}
\end{subfigure}
\caption{(a) Mono-compartment model of the lung. The Bronchial tree is collapsed in the tube having the total resistance $R$ and the alveoli are mimicked by balloons characterized by their elastance $E$. (b) Parallel bi-compartment model. This model better respects the anatomy of the respiratory system.}
\label{fig:=D_model}
\end{figure}

The more tractable model to mimic the lung mechanics and ventilation is to mimic separately its resistive and elastic parts. We represent the resistive tree using a rigid tube with given length $l$ and radius $r$. The resistance $R$ of such a tube can be calculated by using Poiseuille's law that depends on the air viscosity $\mu_\text{air}$\cite{Poiseuille}.
\begin{equation}
    R = \frac{8\mu_\text{air} l}{\pi r^4}
\end{equation}
The elastic part of the lung can be mimicked with an elastic balloon with elastance property $E$.
Figure \ref{fig:MonoCompartment} is a representation of such a model. However, for  this study, the model used is slightly more sophisticated to get a better representation of the distribution of the ventilation, see figure \ref{fig:BiCompartment}. The profile of the pressure used to mimic the muscular action is taken from L. Hao et al. \cite{Hao2019} and represents a standard for tidal breathing. It is necessary to highlight that the hypotheses of linearity used in this model are respected in the regime of tidal breathing \cite{BatesBook}. 

The fundamental equation that links the resistance $R$ and the elastance $E$ to the Pressure $P$, Volume $V$ and its derivative in time $\dot{V}$ can be easily derived from the Mono-compartment model:
\begin{equation}
P_{ext} - P_{alv} = \Delta P(t) = R\dot{V}(t)
\end{equation}
\begin{equation}
P_{alv} - P_{pl} = P_{el}(t) = EV(t)
\end{equation}
$P_{el}$ represents the pressure drop between the acini and the pleural space, this depends on the elastance of the compartment. $\Delta P$ is the air pressure drop between the airways opening and the acini and it takes into account the resistance of the airways and the parenchyma.
The total pressure drop of the model is the sum of the two contributions: 
\begin{equation}
P(t) = P_{el} + \Delta P(t)
\end{equation}This equation holds true regardless of whether $P(t)$ is applied at airways' opening or at the outside of the elastic compartment \cite{BatesBook}.

In the parallel model, figure \ref{fig:BiCompartment}, there are two governing equations, one for each compartment, respectively of volume $V_1$ and $V_2$:
\begin{equation}
\spalignsys{P(t) = E_1V_1(t) + (R_1 + R_t)\dot{V}_1(t) + R_t\dot{V}_2(t);
P(t) = E_2V_2(t) + (R_2 + R_t)\dot{V}_2(t) + R_t\dot{V}_1(t)
\label{eq:1}
}
\end{equation}
where $R_1$ and $R_2$ refers to the resistances of each bronchi, $R_t$ is the resistance of the trachea and $E_1$ and $E_2$ are the elastances of the left and right lung, respectively, see figure \ref{fig:BiCompartment}. These are the parameters of the model. $V_1(t)$ and $V_2(t)$ are the volumes associated to each lung and $P(t)$ is the muscular pressure that drives the lung ventilation.

Let us take the derivative of each equation:
\begin{equation}
\spalignsys{\dot{P}(t) = E_1\dot{V}_1(t) + (R_1 + R_t)\ddot{V}_1(t) + R_t\ddot{V}_2(t);
\dot{P}(t) = E_2\dot{V}_2(t) + (R_2 + R_t)\ddot{V}_2(t) + R_t\ddot{V}_1(t)
\label{eq:2}
}
\end{equation}
We substitute $\ddot{V}_2$ from eq \eqref{eq:2}a into eq \eqref{eq:2}b and replace $\dot{V}_2$ with its expression derived in eq \eqref{eq:1}a. The equation for compartment 1 alone is:
\begin{equation}
\label{eq:partialParallelModel}
\begin{split}
    R_2\dot{P}(t) + E_2P(t) & = \big[R_1R_2 + R_t(R_1 + R_2)\big]\ddot{V}_1(t) + \\ & + \big[(R_2 + R_t)E_1 + (R_1 + R_t)E_2\big]\dot{V}_1(t) + E_1E_2V_1(t)
\end{split}
\end{equation}
Because the model is symmetric, the equation for compartment 2 is the same as \eqref{eq:partialParallelModel} with inverted indexes 1 and 2.
Then remembering that $V(t) = V_1(t) + V_2(t)$ the referral equation for the bi-compartment parallel model is:
\begin{equation}
\label{eq:ParallelModel}
\begin{split}
(R_1 + R_2)\dot{P}(t) + (E_1 + E_2)P(t) & = \big[R_1R_2 +R_t(R_1 + R_2)\big]\ddot{V}(t) + \\ & + \big[(R_2 + R_t)E_1 + (R_1 + R_t)E_2\big]\dot{V}(t) + \\ & + E_1E_2V(t)
\end{split}
\end{equation}
\newpage
\section{Methods}
\subsection{Creation of the dataset}
\label{section:DataSetCreation}
\begin{figure}
\begin{subfigure}{.5\textwidth}
  \centering
  \includegraphics[width=.9\linewidth]{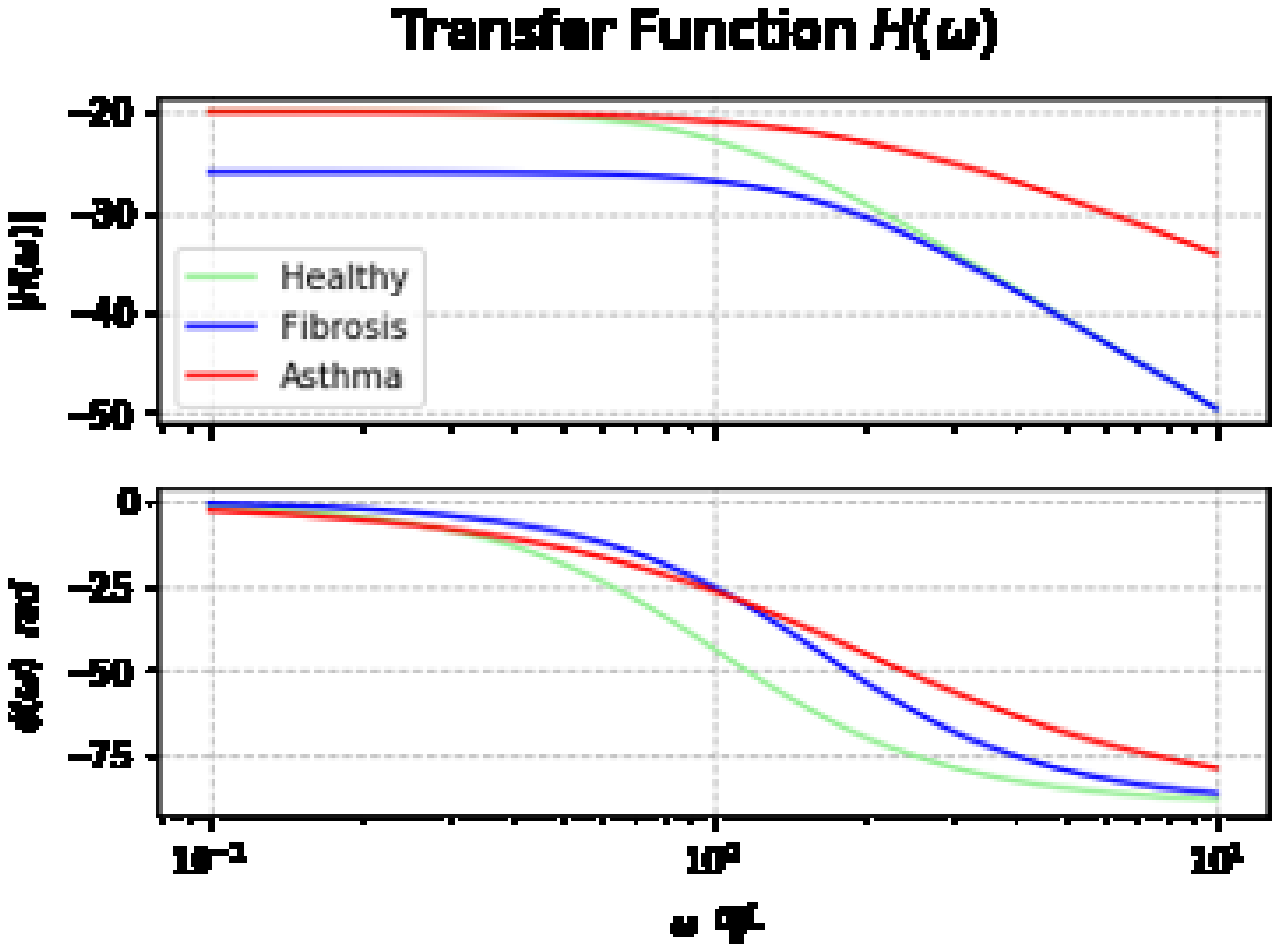}
  \caption{Transfer Function}
  \label{fig:TransferFunction}
\end{subfigure}%
\begin{subfigure}{.5\textwidth}
  \includegraphics[width=.9\linewidth]{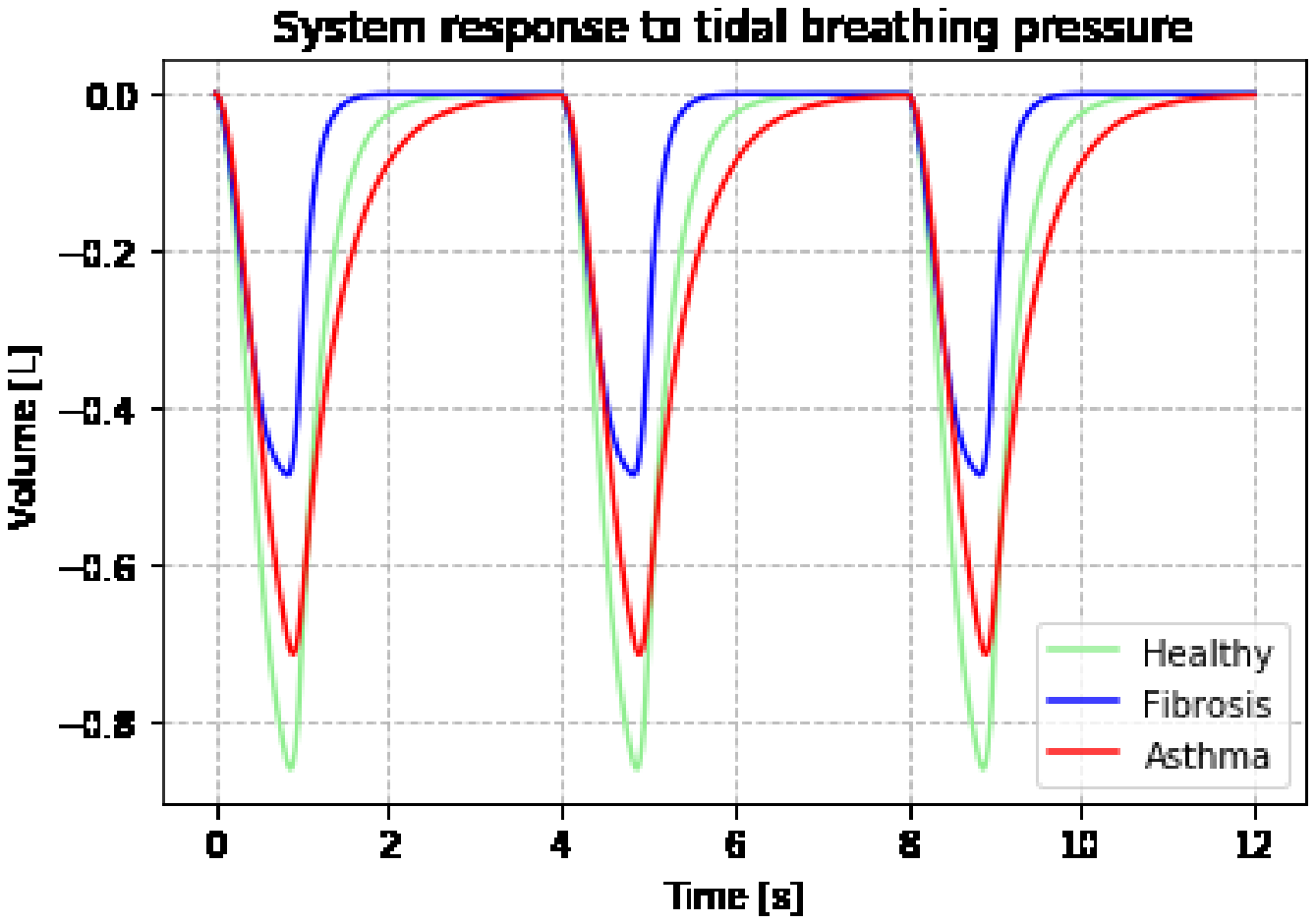}
  \caption{System Response}
  \label{fig:SystemResponse}
\end{subfigure}
\caption{(a) Three different transfer functions, in the subplot above there is the module of the transfer function: $|H(\omega)|$ and below the phase: $\phi(\omega)$.  Increasing the rigidity affects the response of the system at lower frequencies whereas increasing the total resistance affects higher frequencies. Tidal breathing happens at around 0.25 Hz being in the middle of the cutting frequence of $H(\omega)$. Consequently the output of the Volume changes. Figure (b) represents the output of the system (Volumetric signal) for one sample for each class.}
\end{figure}

It is possible to mimic the behavior of healthy subjects by setting physiological values of $R_{eq} = R_t + \frac{R_1R_2}{R_1+R_2}$ and $E_{eq}=\frac{E_1E_2}{E_1+E_2}$. In the literature, they are set to: $R_{eq}=3$ $cmH_2O/L/s$ and $E_{eq} = 10$ $cmH_2O/L$ \cite{Hao2019}.  In this work, we mimic cystic fibrosis by doubling the healthy elastance (doubling the rigidity of the balloons): $E_{eq} = 20$ $cmH_2O/L$, and asthmatic subjects by setting $R_{eq} = 5$ $cmH_2O/L/s$.
It is possible to follow the characteristic approach of electrical analysis \cite{Otis56} in which complex differential equations are studied in the frequency domain through the Laplace transform. The Laplace transform of eq \eqref{eq:ParallelModel} is:
\begin{equation}
    H(s) = \frac{s(R_1+R_2)+(E_1+E_2)}{s^2\Big[R_1R_2 +R_t(R_1 + R_2)\Big] + s\Big[(R_2 + R_t)E_1 + (R_1 + R_t)E_2\Big] + E_1E_2}
    \label{eq:Laplace}
\end{equation}
Figure \ref{fig:TransferFunction} shows the module and phase of the transfer function of the system in the cases of healthy, fibrosis and asthma, while Figure \ref{fig:SystemResponse} shows the responses of each system to physiological $P(t)$.

\begin{figure}
\centering
\begin{subfigure}{.5\textwidth}
  \centering
  \includegraphics[width=.9\linewidth]{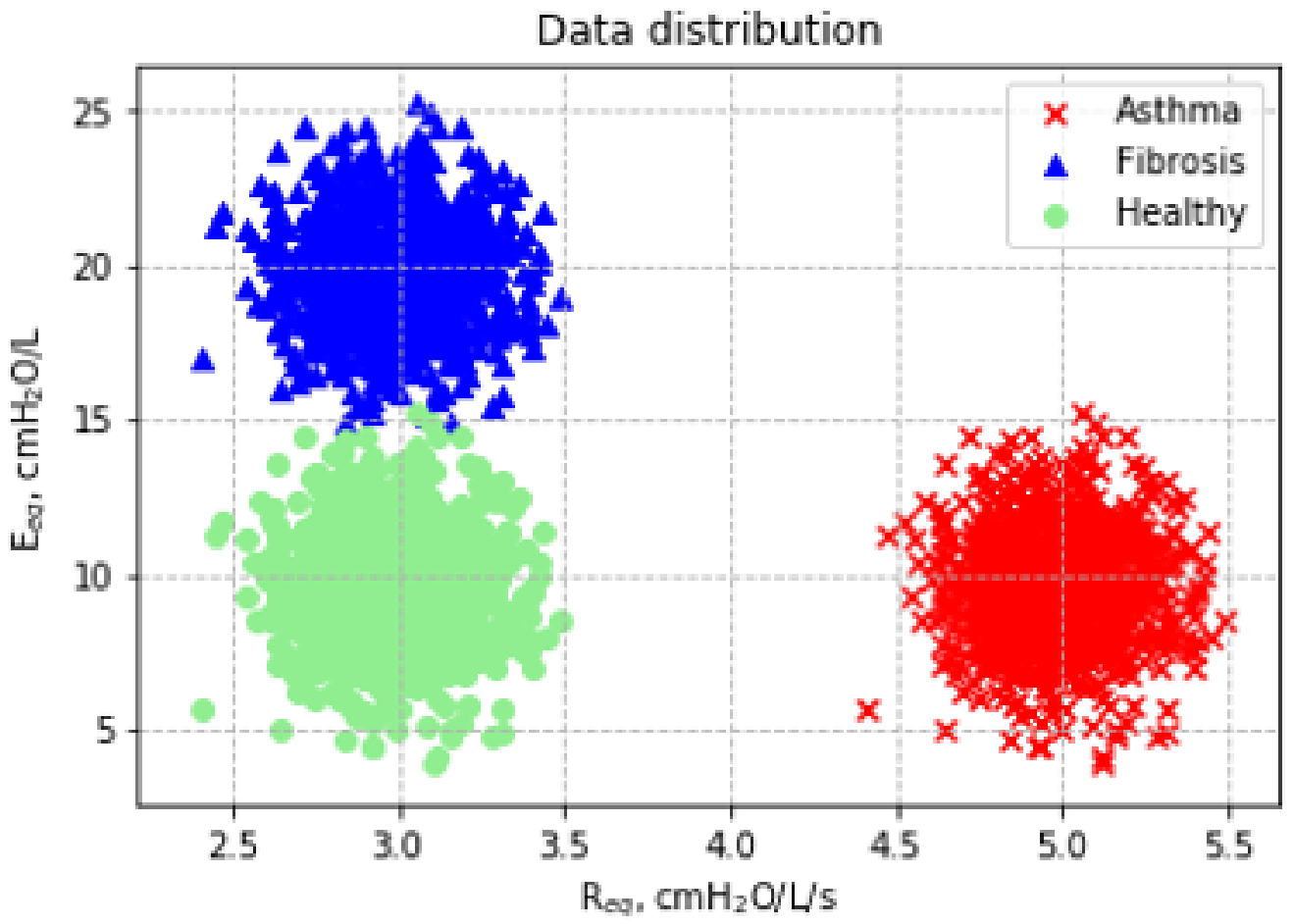}
  \caption{Parameter distribution}
  \label{fig:DataDistribution}
\end{subfigure}%
\begin{subfigure}{.5\textwidth}
  \centering
  \includegraphics[width=.9\linewidth]{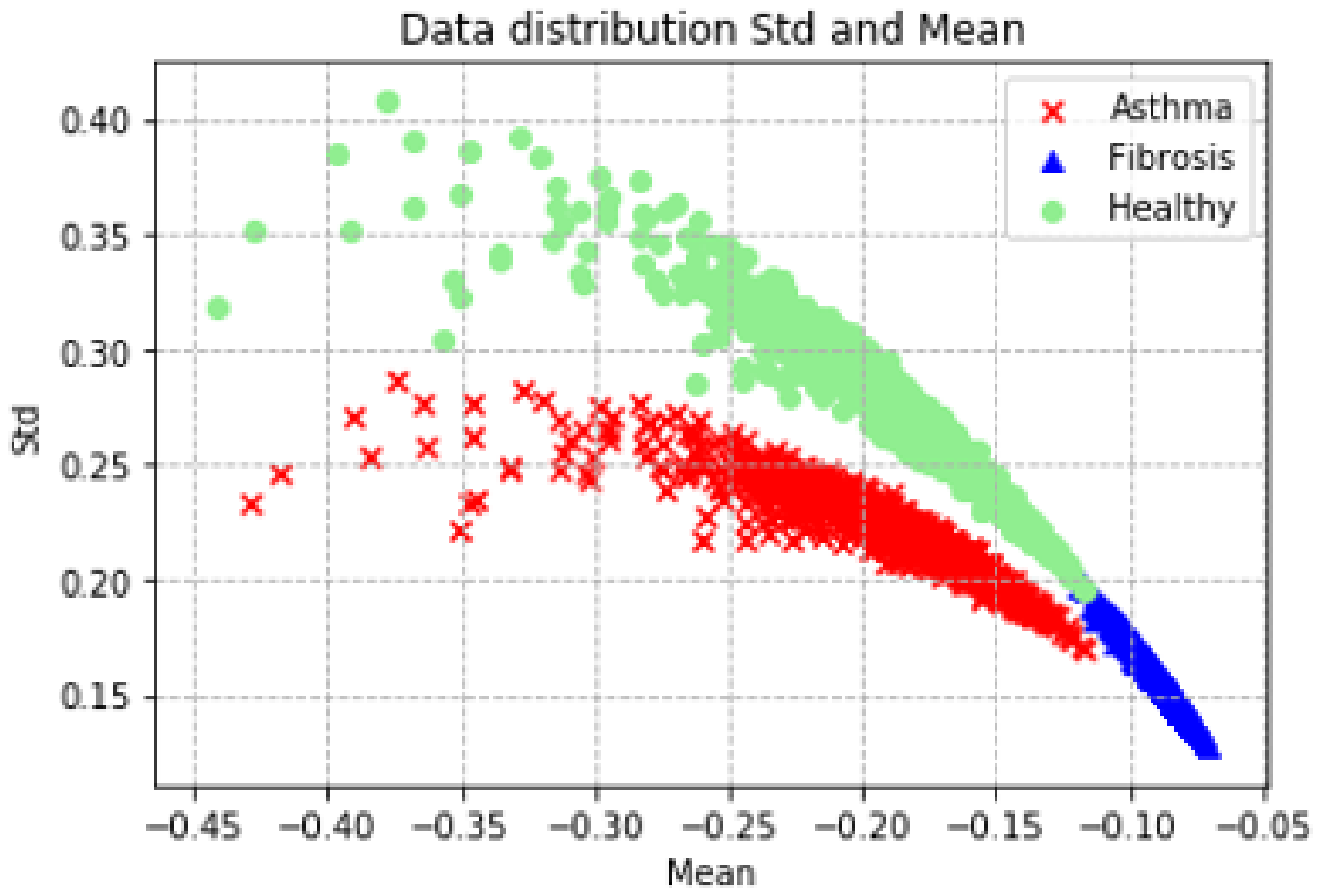}
  \caption{Tidal breathing simulation.}
  \label{fig:DDMeanAndStd}
\end{subfigure}
\caption{(a) Synthetic data distribution, the three different clusters are well visible in this space. (b) Mean and Std features taken from volumetric signals of tidal breathing. These signals are the output of the model as explained in equation \eqref{eq:ParallelModel}}
\end{figure}

Gaussian noise with mean $\mu = 0$ and standard deviation $\sigma = 0.5$ for the $R_{eq}$ parameter and $\sigma = 5$ for the $E_{eq}$ parameter, is added to $R_{eq}$ and $E_{eq}$ to mimic physiological diversity among different subjects as showed in figure \ref{fig:DataDistribution}, there are 1000 samples for each class.

\subsection{Training Machine Learning algorithms}
Before talking about the training, a short summary for each classifier used is reported. The implementation has been done using Python and the open-source library scikit-learn.
\subsubsection{Naive Bayes}
Naive Bayes classifier is a classifier that naively apply the Bayes theorem. A classifier is a function $f$ that take an example $\pmb{x} = (x_1, x_2, ..., x_n)$ where $x_i$ is the $i^{th}$ feature and transform it in a class $y$. According to Bayes theorem the probability $P$ of an example $\pmb{x}$ being class $y$ is:
\begin{equation}
    P(y\mid \pmb{x})=\frac{P(\pmb{x}\mid y)P(y)}{P(\pmb{x})}
\end{equation}
Assuming that all the attributes are independent, the likelihood is:
\begin{equation}
    P(\pmb{x}\mid y)=P(x_1, x_2, ..., x_n \mid y) = \prod_{i=1}^{n} P(x_i\mid y),
\end{equation}
hence, rewriting the posterior probability:
\begin{equation}
    \label{eq:eq12}
    P(y\mid\pmb{x}) = \frac{P(y)\prod^n_{i=1}P(x_i\mid y)}{P(\pmb{x})}
\end{equation}
because $P(\pmb{x})$ is constant with regards of $y$, equation \ref{eq:eq12} can be rewritten and used to define the Naive Bayes (NB) classifier:

\begin{align}\begin{aligned}P(y \mid \pmb{x}) \propto P(y) \prod_{i=1}^{n} P(x_i \mid y)\\\Downarrow\\\hat{y} = \arg\max_y P(y) \prod_{i=1}^{n} P(x_i \mid y),\end{aligned}\end{align}   

Albeit the hypothesis of independence among attributes is never respected in real world, this classifier still has very good performance. Indeed, it has been observed that its classification accuracy is not determined by the dependencies but rather by the distribution of dependencies among all attributes over classes \cite{Zhang,Domingos1997}.
\subsubsection{Logistic Regression}
Despite its name, this is actually a classification algorithm. This is a linear classifier used normally for binary classification even though it can be extended to multiclass through different techniques like OvR (One versus Rest) or multinomial \cite{Raschka2018}.
In our work, the \textit{newton-cg} solver has been used together with \textit{multinomial} multiclass. In this configuration, the $\ell_2$  regularization is used and the solver learns a true multinomial logistic regression model using the cross-entropy loss function \cite{Schulz2011}. Using these settings allows the estimated probabilities to be better calibrated than the default "one-vs-rest" setting, as suggested in the official documentation of scikit-learn \cite{scikitLogReg}.

\comment{Logistic regression models the posterior probability for binary classification as a sigmoid function:
\begin{equation}
    P(y = \pm 1 \mid \pmb{x}) \equiv \frac{1}{1 + e^{y\pmb{w}^T\pmb{x}}}
\end{equation}
where $\pmb{w} \in \mathbb{R}^n$ is the vector of trainable weights, $\pmb{x}$ is the data and $y$ is the class label.
For 2 classes training data, given \textcolor{blue}{$\ell_2$ penalization} the loss function to minimize is the cross-entropy which corresponds with the maximization of the likelihood. The regularized negative log-likelihood to minimize is:
\begin{equation}
    \min_{\pmb{w}}\Big( \textcolor{blue}{\frac{1}{2}\pmb{w}^T\pmb{w}} + C \sum_{i=1}^n \log\big(1 + e^{- y_i (w_i^Tx_i) }\big)\Big) .
\end{equation}

where $C > 0$ is the penalty parameter which is set to 1 by default in scikit-learn.} 

When multinomial multiclass is used, the posterior probabilities are given by a softmax transformation of linear functions of the feature variables \cite{Schulz2011}:
\begin{equation}
    P(y_k|\pmb{x}) = \frac{e^{\pmb{w}^T_k\pmb{x}}}{\sum_je^{w^T_j\pmb{x}}}
\end{equation}
where $\pmb{w} \in \mathbb{R}^n$ is the vector of trainable weights, $\pmb{x}$ is the feature vector and $y$ is the class label.
Using 1-of-K encoding scheme, it is possible to define a matrix $\pmb{T}$ composed by $n$ rows (being N the total number of features for each class) and $k$ columns (being K the total number of classes) \cite{Schulz2011}. In our case N=2 and K=3. Each vector $\pmb{t_n}$ will have one in the position of its class and zeros all over the rest.
In this scenario, the \textit{cross-entropy} loss function to minimize for the multinomial classification regularized with $\ell_2$  is: 
\begin{equation}
    \min_{\pmb{w}}\Big(\frac{1}{2}\pmb{w}^T\pmb{w} - \sum^N_{n=1}\sum^K_{k=1}t_{nk}\ln(\hat{y}_{nk})\Big)
\end{equation}
being $\hat{y}_{nk} = P(y_k|\pmb{x_n})$.
\subsubsection{Perceptron}
For linear separable datasets, Perceptron can achieve perfect performances because it guarantees to find a solution, hence the learning rate $\eta$ is not essential and by default is set to 1.0 in scikit-learn. In our implementation, the loss function is the number of mislabelled samples and it is not regularized. The weights of the model are updated on mistakes as follows:
\begin{equation}
    w_{j+1} = w_j + \eta\big(y^{(i)} - \hat{y}^{(i)}\big)x^{(i)}_j
\end{equation}
where $i$ is the sample, $j$ is the feature, $y$ is the target and $\hat{y}$ its respective prediction. \cite{Raschka2018}.
The weights are updated with Stochastic Gradient Descent (SGD) optimizer, meaning that the gradient of the loss is estimated for each sample at a time and the model is updated along the way \cite{scikitPerceptron}.

\subsubsection{Support-Vector Machines}
\begin{figure}[H]
\centering
\includegraphics[width=.5\textwidth]{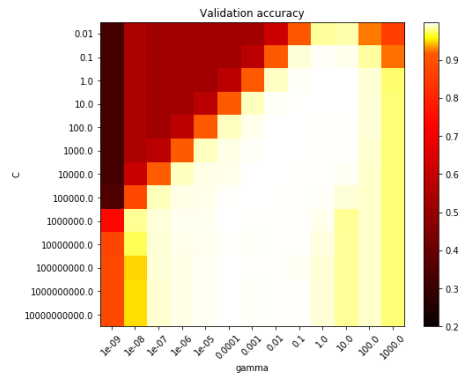}
\caption{Heat map for setting the best pair of $\gamma$ and $C$ in rbf kernel function for SVMs, brightest colors correspond to highest validation accuracy.}
\label{fig:rbfKernel}
\end{figure}

This is one of the most robust supervised ML algorithms, it is used for both regression and classification problems and it can be used in a non-linear fashion thanks to kernel tricks \cite{hofmann2008}. In SVMs, we used the Radial Basis Function (rbf) kernel:
\begin{equation}
    K(X,X')= e^{\gamma||X-X'||^2}
\end{equation}
When implementing this function, there are 2 parameters required:
\begin{itemize}
\item $\gamma$, which is the term in the expression of the rbf and it is the coefficient of multiplication for the euclidean distance
\item $C$, which is the error Cost, this is not directly related to the kernel function, instead it is the penalty associated with misclassified instances.
\end{itemize}
Setting these parameters together is important for achieving good results. In our case, a vast selection of pairs reports similar results, as shown in figure \ref{fig:rbfKernel}.
\subsubsection{Random Forest}
Random Forest (RF) is an ensemble of decision tree estimators in which each estimator classifier has been used with 100 trees and Gini impurity as criterion. The class prediction is performed by averaging the probabilistic prediction of each estimator instead of using the voting system as implemented in its original publication \cite{Breiman2001}. Random Forests follows the exhaustive search approach for the construction of each tree, where the main steps are listed in algorithm \ref{alg:PseudoCode_DT} \cite{DiDio19}. 
\comment{
The used criterion stop used is the Gini impurity 
\begin{equation}
    \displaystyle \operatorname {I} _{G}(p)=\sum _{i=1}^{J}p_{i}\sum _{k\neq i}p_{k}
    =1-\sum _{i=1}^{J}{p_{i}}^{2} \ , 
\end{equation}
which is an indicator of how often an item chosen randomly from the subset is mislabeled if it was labeled in a random way according to the distribution of labels in the subset \cite{DiDio19}.}

\begin{algorithm}[h]
\caption{Pseudocode for tree construction - \textbf{Exhaustive search}}\label{alg:PseudoCode_DT}
\begin{algorithmic}[1]
\State $\text{Start at the root node}$
\For{each $X$}
\State \Longunderstack[l]{Find the set $S$ that minimizes the sum of the node impurities \\ in the two child nodes and choose the split $S^\ast \in X^\ast$ that gives the \\ minimum overall $X$ and $S$}
\EndFor
\If{Stopping criterion is reached}
    \State{Exit}
\Else
    \State{Apply step 2 to each child node in turn}
\EndIf
\end{algorithmic}
\end{algorithm}
\subsection{Training} \label{subsection:Training}
Once the dataset is ready, simulations are performed and significant statistical features are extracted from the signals (examples of output signals of the system are shown in Figure \ref{fig:SystemResponse}). Here, for facilitating the graphical representation, two features are extracted: mean and standard deviation. These features have enough information to correctly differentiate among the three classes. Before training each of the previous models, standard scaling has been fit on the training set and applied on both training and test sets. 
The dataset has been split randomly by keeping the size of the training set at 80\% of the total dataset. Hence performances have been evaluated on 800 samples after having checked the correct balance among the classes. The classifiers previously reported have been trained and tested and their relative decision plots can be seen in figure \ref{fig:Classifiers}.
\section{Results} \label{section:Results}
\subsection{Lung model}
\label{LungModel}
\begin{figure}
\centering
\begin{subfigure}{.5\textwidth}
  \centering
  \includegraphics[width=\linewidth]{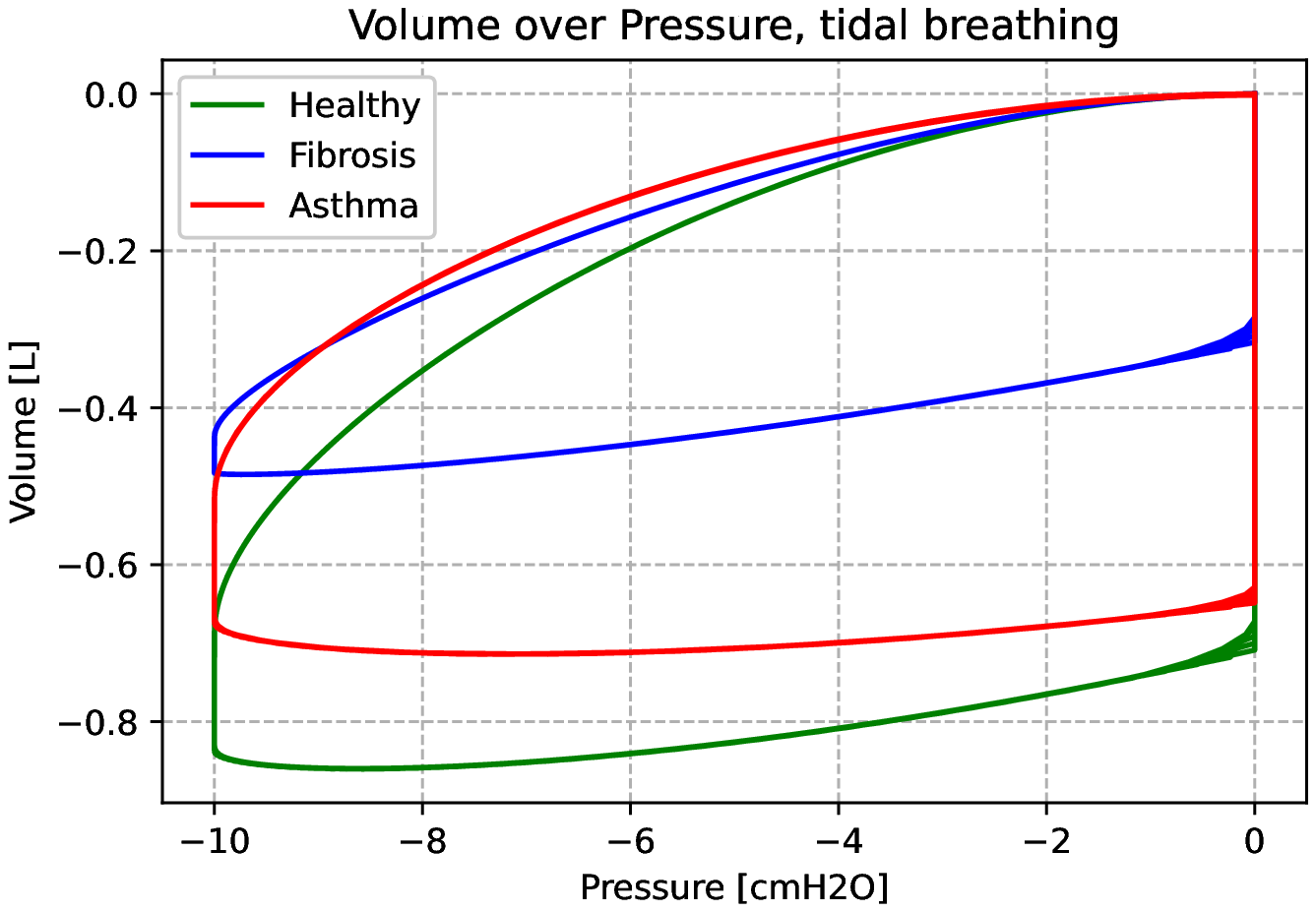}
  \caption{Pressure-Volume loop}
  \label{fig:PVplot}
\end{subfigure}%
\begin{subfigure}{.5\textwidth}
  \centering
  \includegraphics[width=\linewidth]{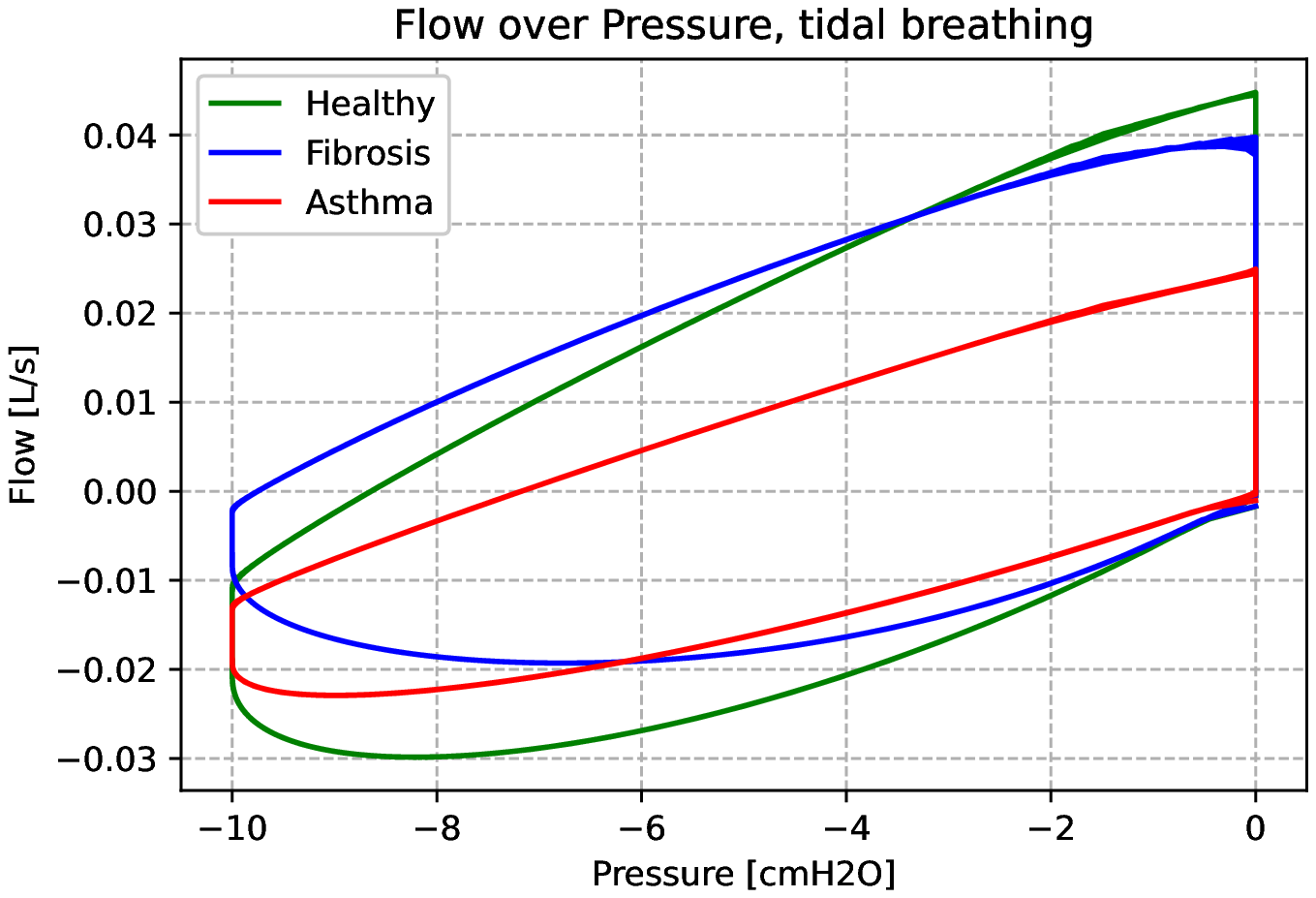}
  \caption{Flow-Volume loop}
  \label{fig:FVplot}
\end{subfigure}
\caption{}
\label{fig:PVandFV}
\end{figure}

The parallel model used is a good representation of the lung when detailed geometrical characteristics are not important to model. Working with this model allows to control the resistance and elastance of the respiratory system, allowing the simulation of certain diseases. However, it is important to ensure that the results given by our model are coherent with reality. Because of this, the output signals of volumes and flows have been observed and visually compared with real signals, see figure \ref{fig:SystemResponse}. The flow $\Phi$ has been calculated as $\Phi(t) = \partial V(t) / \partial t$. Pressure-Volume plots and Flow-Volume plots have been evaluated for each class, see figure \ref{fig:PVandFV}.
In typical Pressure-Volume plots a decreasing of compliance is manifested as a shift on the right of the loop. However the model that we are using is pressure driven. In a pressure driven model, when the compliance decreases, the pressure control yields less and less volume for the same pressure level, causing a lowering of the loop curve.  Asthmatic subjects on the other hand are simulated as having same compliance as healthy subjects but greater resistance, and because of this, their flow is lower than the others in a pressure driven model as visible in Figure \ref{fig:FVplot}.

\begin{figure}
\centering
\begin{subfigure}{.5\textwidth}
  \centering
  \includegraphics[width=\linewidth]{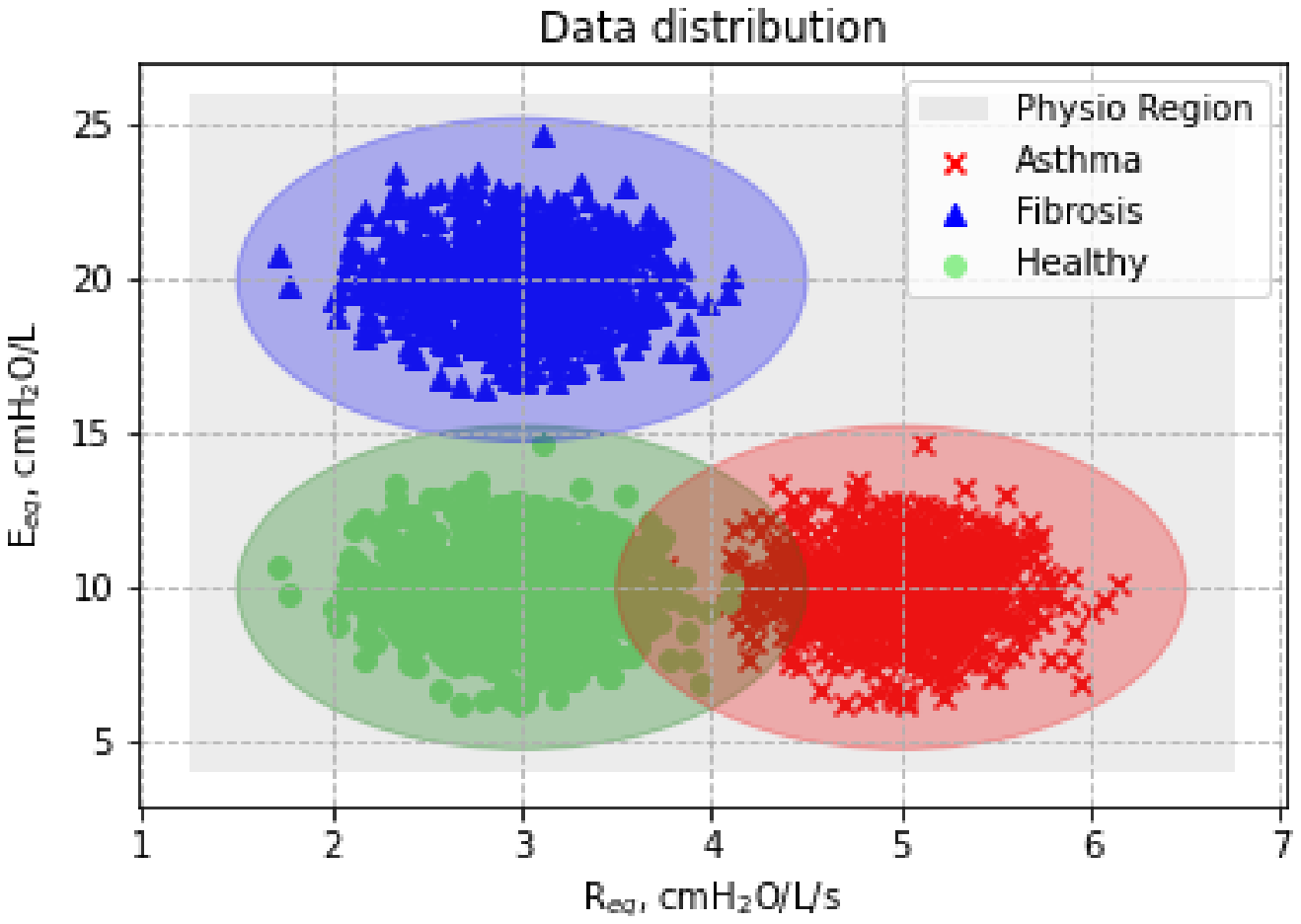}
  \caption{}
  \label{fig:EllRegions}
\end{subfigure}%
\begin{subfigure}{.5\textwidth}
  \centering
  \includegraphics[width=\linewidth]{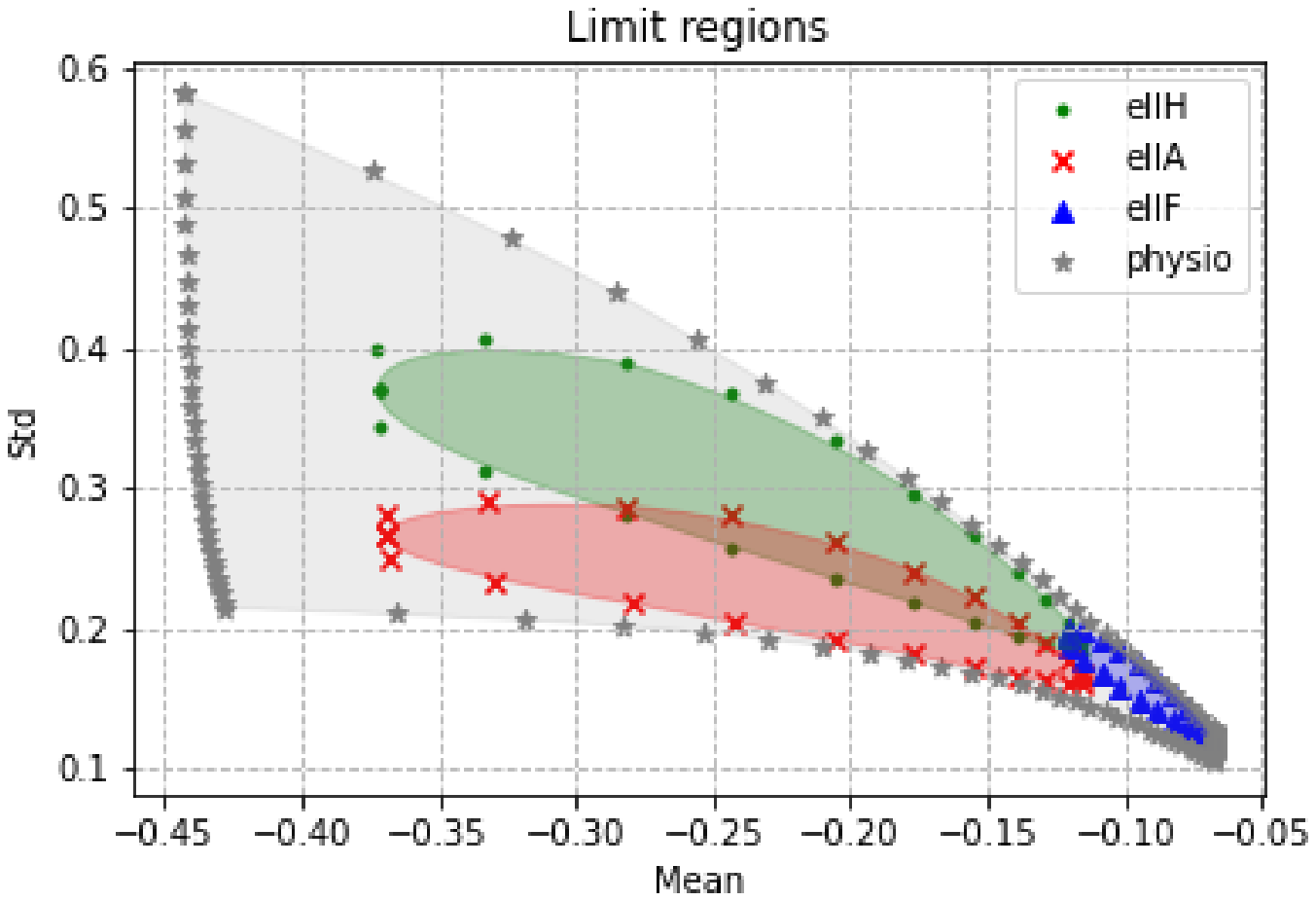}
  \caption{}
  \label{fig:EllOutRegions}
\end{subfigure}
\caption{Data distribution and visual representation of each set in the $(R_{eq}, E_{eq})$ space (a) and in the $(\mu, \sigma)$ space (b)}
\label{fig:EllComplete}
\end{figure}

{
The model performs a transformation $(R_{eq}, E_{eq}) \in \mathbb{R}^2 \rightarrow \mathbb{S}$, with $\mathbb{S} \subseteq \mathbb{R}$, which is the signal space, in our case the signal is $V(t) \in \mathbb{S}$. By extracting features from the signal space, we pass to $\mathbb{R}^N$, where $N$ is the number of features extracted. Here, we have extracted two features, therefore we have a mapping $\mathbb{R}^2: (R_{eq}, E_{eq}) \rightarrow \mathbb{R}^2: (\mu, \sigma)$ where $\mu$ and $\sigma$ are the extracted features, respectively the mean and the standard deviation of the signal. The image of this mapping is particularly useful to set limits in the prediction of the AI. Indeed, for measurements that are not contained in the image, and thus inconsistent, it makes sense to have a separate treatment 
that will alert when a measurement must be discarded and replaced by a new one because it is not physiological.
To achieve this, a physiological region has been defined (gray area in Figure \ref{fig:EllRegions}), the boundary of the rectangle region (the physiological set) is passed to the system and the output path is then patched again to form a polygon of acceptable measurements (gray area in figure \ref{fig:EllOutRegions}). Out of this patch of acceptable measurements, the data will not be passed to the AI for prediction and a message of "wrong acquisition" will be displayed.}
{In Figure \ref{fig:EllComplete}, the data distribution is obtained with standard deviation $\sigma(R_{eq}) = 1$ $cmH_2O/L/s$ and $\sigma(E_{eq}) = 3.5$ $cmH_2O/L$ to allow for a better spreading and overlapping. Moreover, because we want the elliptic patch of each class to be comprehensive of almost all its possible samples, we fix the width and the height of the ellipse to be three times the respective standard deviation.}

\label{subsec:LungResults}
\subsection{Machine Learning results}
\label{subsec:ClassifierResults}
\begin{figure}[h]
\includegraphics[width=\linewidth]{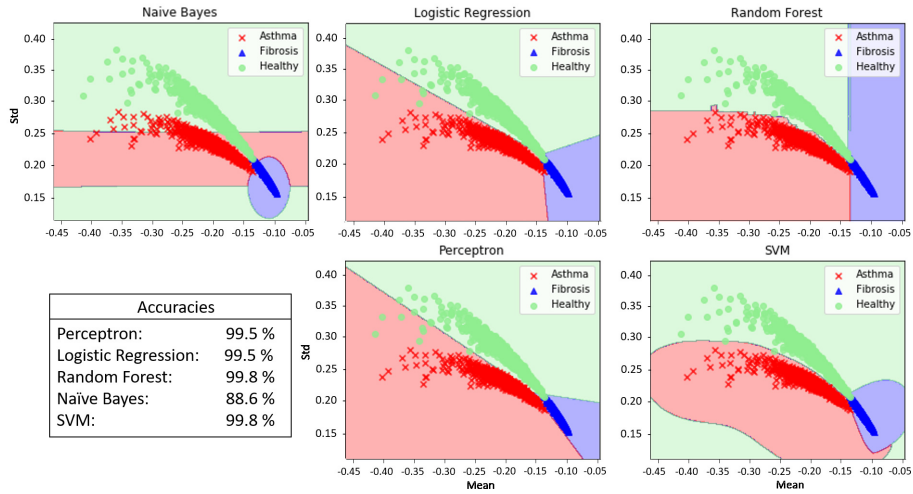}
\caption{Decision regions and accuracy of the implemented classifiers. The regions should be shortened according to the limitations indicated in figure \ref{fig:EllOutRegions}.}
\label{fig:Classifiers}
\end{figure}
Multiple Machine Learning algorithms have been tested against synthetic data retrievable using spirometers. Their performances are shown in figure \ref{fig:Classifiers} in terms of accuracy and decision boundaries. In this section, there are some considerations regarding the different classifiers used. By looking at the data distribution in figure \ref{fig:DDMeanAndStd}, we observe that a linear separation will not be perfect but still good. Multinomial logistic regression and Perceptron have been used and compared for such a linear separation. Their performances are then compared with other commonly used non-linear classifiers; the first to be tried has been Naive Bayes because of its interpretability and ease of use, however its poor performance led us to try more sophisticated models like SVM with rbf kernel and RF. These latter classifiers lead to great performance, even if they have local errors close to the decision region boundaries. Nevertheless, these regions are characterized by spurious areas located on both bottom corners in the case of SVM (figure \ref{fig:Classifiers} SVM subplot) and the right-up region in the case of RF.
\begin{itemize}
\item \textbf{Naive Bayes}: It is possible to see how this algorithm is not suited for this kind of multi-class classification. Indeed, this classifier is usually used for binary classification (it is used a lot for spam detection). In this particular dataset, Naive Bayes fails in detecting a border between healthy and asthmatic subjects and the found boarder is not significant compared with other classifiers.
\item \textbf{Logistic Regression}: This classifier is one of the most used in biomedical applications, both for its easy comprehension and its great performances when the classes are linearly separable. In this case, classes are not linearly separable. Nevertheless, this keeps a good level of generalization without renouncing to ease of usage, comprehension and good performances.
\item \textbf{Random Forest}: RF is probably one of the most powerful classifiers. It is used also in biomedical applications for its good performances and its resistance to overfitting. It is an ensemble method where multiple decision trees (DTs) are singularly trained. Finally the average of the predictions of all the estimators will be used to make the decision of the RF classifier. DTs are used in medicine because of their clarity in the decision, however with RF there is a loss of this explainability in the decision, caused by an enhancing of complexity due to the ensemble.
\item \textbf{Perceptron}: This is a linear model as long as only one layer is provided. It is powerful and performs very well in this particular situation. It can be useful to increase the deepness of its structure once there are lots of features and their relationships are not of easy interpretation. Also, in contrast with logistic regression it can be easily used for non-linearity. 
\item \textbf{Support Vector Machine}: This model is widely used in a lot of applications. In this work, the Radial Basic Function kernel has been used. Its non-linear nature allows to follow better the separation between healthy and asthma. In this dataset, this is the most performant model to distinguish these two classes. However, like the Naive Bayes it creates a green area below the red and blue zone that are incorrect.
\end{itemize}
In contrast with Deep Learning, ML models are normally faster to train. Figure \ref{fig:TrainTime} shows the differences of training times among the used classifiers. Even if the timings are very small, it is interesting to see, for instance, how fast the Perceptron is compared to Random Forest (RF). This is an interesting property for large datasets.
\comment{\begin{figure}
	\centering
    \includegraphics[width=.6\textwidth]{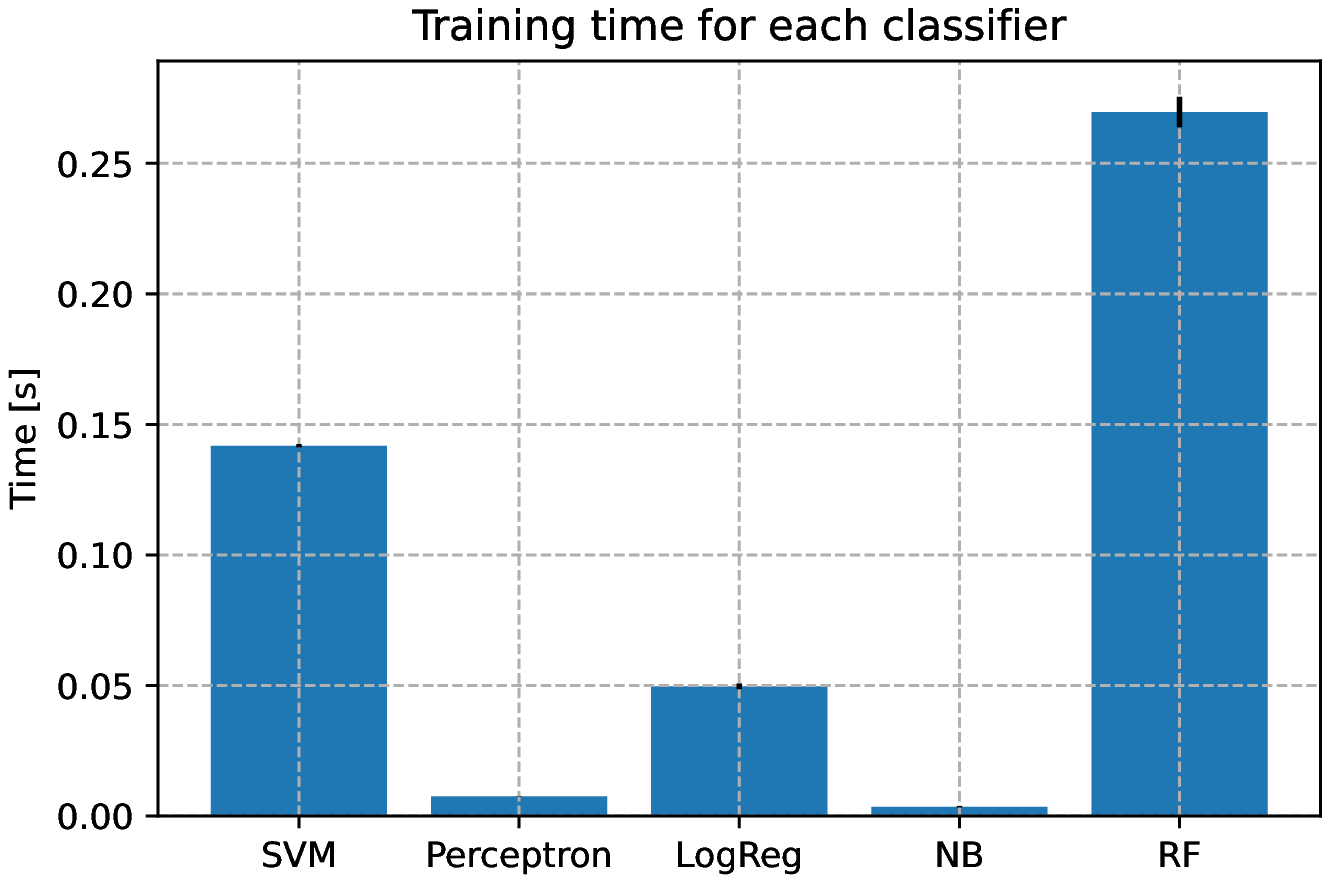}
    \caption{Different timings of training for each classifiers. Trainings are performed using the CPU runtime on Colab}
    \label{fig:TrainTime}
\end{figure}}
Figure \ref{fig:ROC} shows the Receiver Operating Characteristic (ROC) curves and the respective calculated Area Under the Curve (AUC) for each classifier. ROC curves have been adapted for multiclass using the macro averaging technique. As expected, SVM and RF outperforms the other classifiers, however, using this metric is possible to observe the difference between Logistic Regression and Perceptron. The origin of this difference is probably on the separation between asthma and fibrosis as observable in figure \ref{fig:Classifiers}.

\begin{figure}
\centering
\begin{subfigure}{.45\textwidth}
  	\centering
    \includegraphics[width=.99\textwidth]{TrainingTime.eps}
    \caption{}
    \label{fig:TrainTime}
\end{subfigure}
\begin{subfigure}{.45\textwidth}
    \centering
    \includegraphics[width=.99\textwidth]{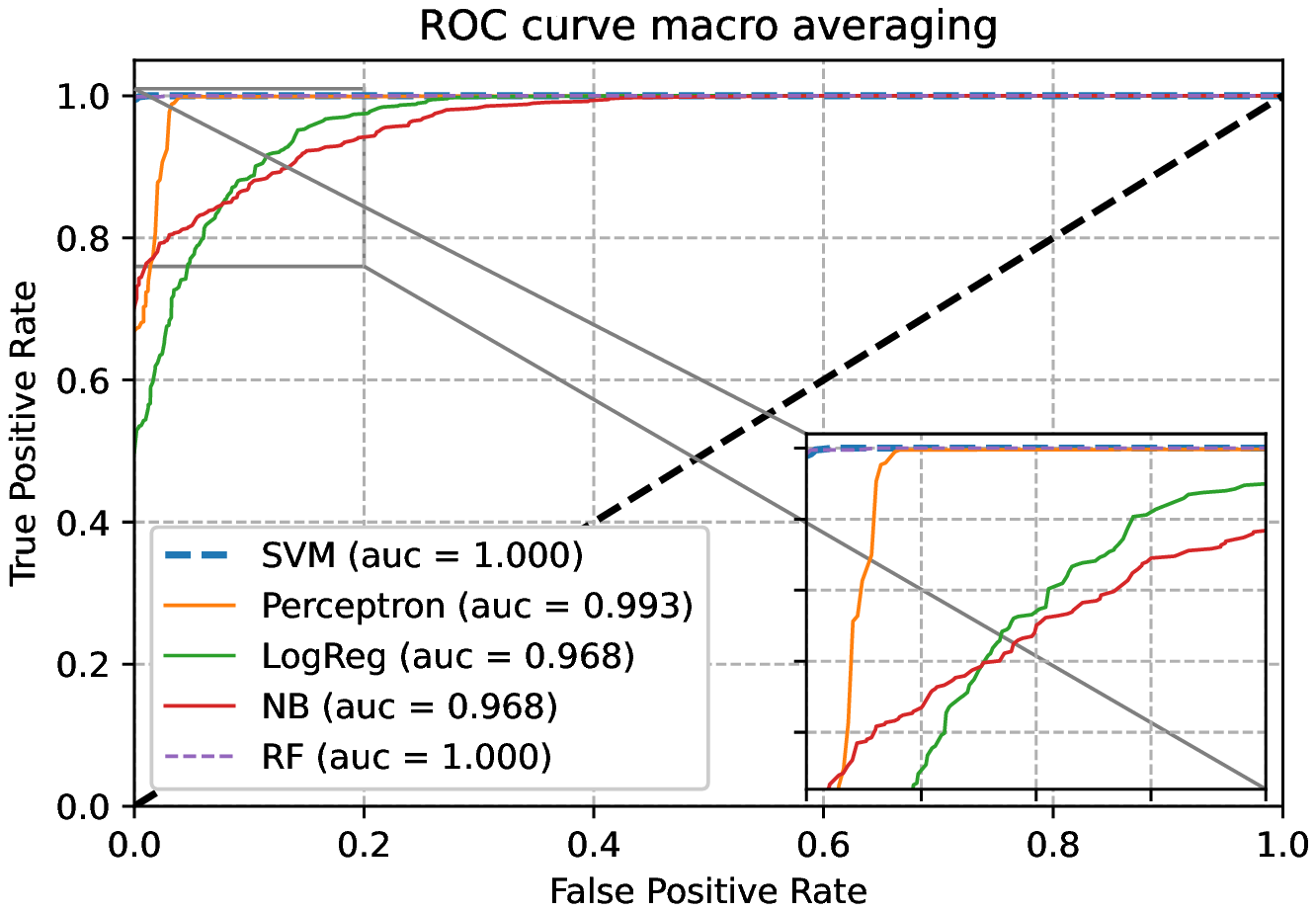}
    \caption{}
    \label{fig:ROC}
\end{subfigure}%
\caption{(a) Different timings of training for each classifiers. Trainings are performed using the CPU runtime on Colab. (b) Macro averaging of Receiver Operating Characteristic curve for each curve, zoom on the upper left part. RF and SVM are overlapped and their value is fixed on 1.}
\end{figure}

\comment{
\begin{figure}
\centering
\includegraphics[width=.6\textwidth]{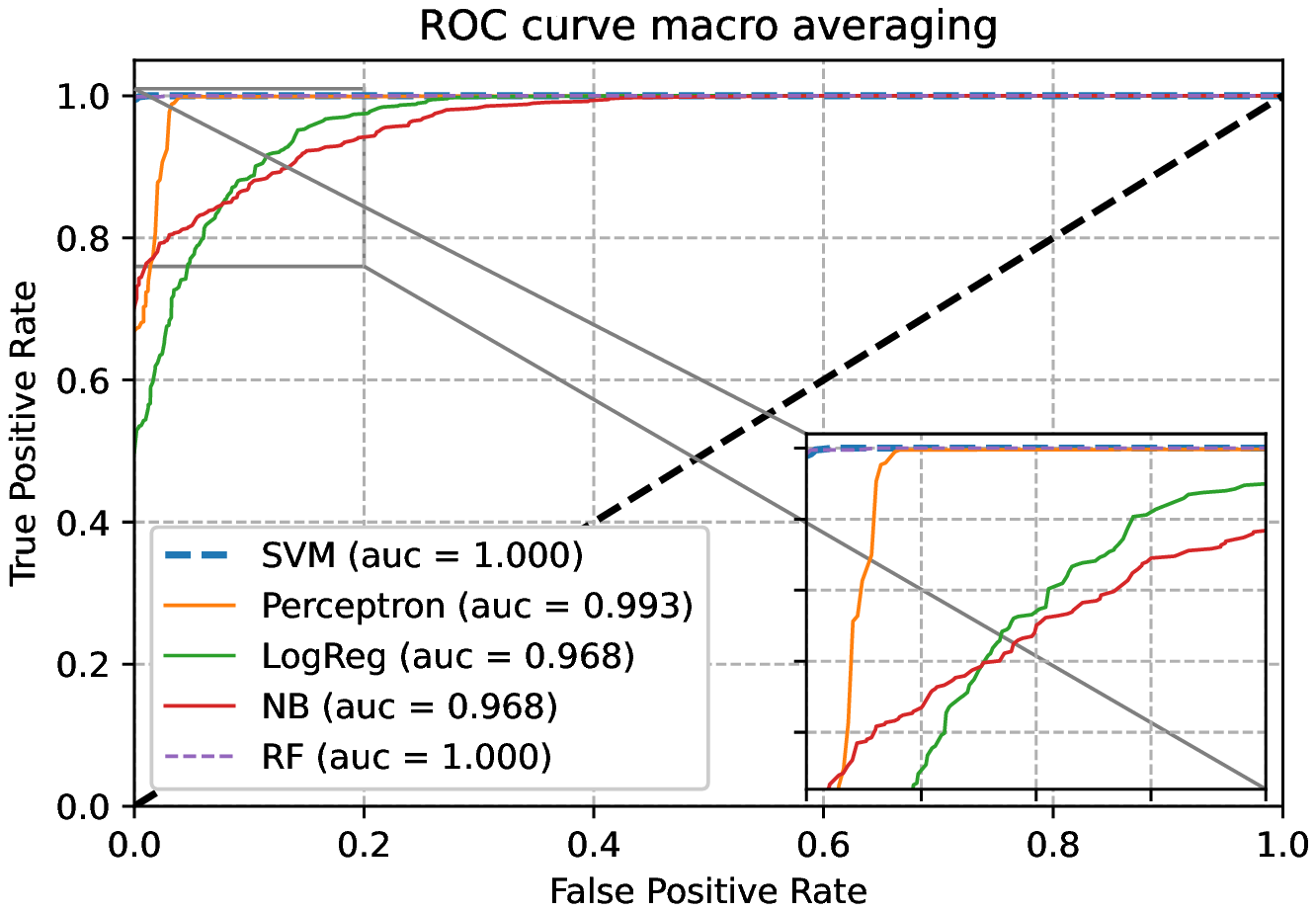}
\caption{Macro averaging of Receiver Operating Characteristic curve for each curve, zoom on the upper left part. RF and SVM are overlapped and their value is fixed on 1.}
\label{fig:ROC}
\end{figure}}

\par It is worth to mention that we carried out supplementary simulations using data more spread than in figure \ref{fig:DataDistribution}, which resulted in even more overlaps between the different classes (see figure \ref{fig:EllComplete}). In these cases, the decision regions of each classifier stay very similar to those shown in figure \ref{fig:Classifiers} with a resulting lower accuracy due to the overlapping.

\newpage
\section{Conclusions and outlook}
\label{section:Discussion}
As a brief recap, in this work the following has been done:

A second order ODE mathematical model of the lung has been used to generate synthetic data of asthma, cystic fibrosis and healthy subjects. This data has been used to train Machine Learning models. The models have been evaluated on different synthetic data sampled from the same distribution of the training set. A solution has been proposed for non physiological measurements, see subsection \ref{subsec:LungResults}. Finally, differences among the classifiers have been studied in terms of accuracy, ROC curves and training timings.

We elaborated on the potential use of modern ML techniques to diagnose diseases of the human respiratory system. 
The direct conclusion of this work is the ability of ML algorithms to distinguish among linear separable clusters in the $\mathbb{R}^2$  ($R_{eq}$, $E_{eq}$) space, also in the non-linear feature space of $f(g(R_{eq}, E_{eq}))$ where $g: \mathbb{R}^2 \rightarrow \mathbb{S}$ is the parallel lung model and $f: \mathbb{S} \rightarrow \mathbb{R}^N$ is the feature extraction with $N$ being the number of features.

One limitation of our work is the training and testing based entirely on simulated data and the utilization of the same pressure profile for all the classes. After the present proof-of-concept, future work will include training on simulated data and testing on acquired real data. Moreover, the usage of depth camera will be investigated to extract tidal breathing patterns \cite{Wang20}.
Furthermore, the pressure was assumed to be uniform throughout the lungs and there was no difference in the application of the pressure among the three considered cases. This condition is in principle not respected because some patients can increase their muscle effort in order to keep a satisfactory ventilation. However, it is possible to assist the patients and train them to follow a specific pattern while breathing in the spirometer.

To conclude, the different ML models presented are proven in principle reliable, therefore they could provide the physicians with real-time help for the diagnosis decision.

\section*{Acknowledgements}
This project has received funding from the European Union’s Horizon 2020 research and innovation programme under the Marie Curie grant agreement No 847581 and is co-funded by the Région SUD Provence-Alpes-Côte d'Azur and IDEX UCA JEDI.
\begin{figure}[H]
\centering
\begin{subfigure}{.3\textwidth}
  	\centering
    \includegraphics[width=.99\textwidth]{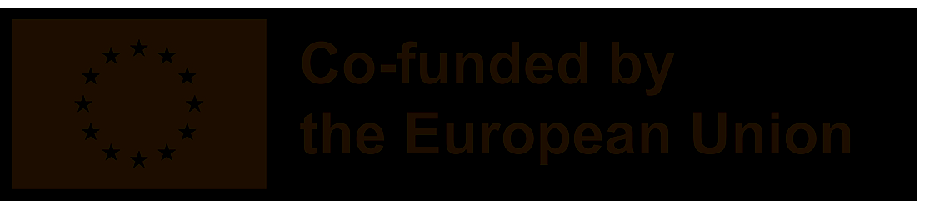}
\end{subfigure}
\begin{subfigure}{.3\textwidth}
    \centering
    \includegraphics[width=.99\textwidth]{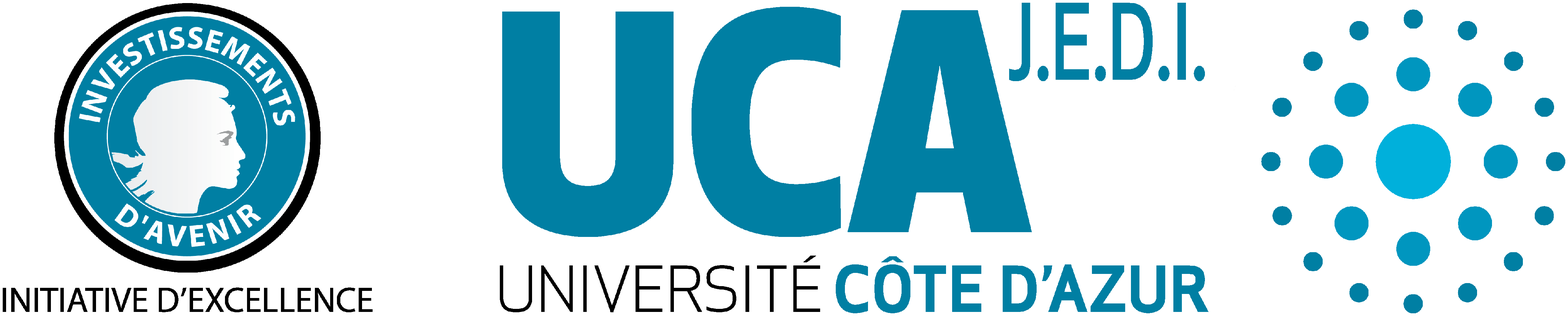}
\end{subfigure}
\begin{subfigure}{.3\textwidth}
    \centering
    \includegraphics[width=.59\textwidth]{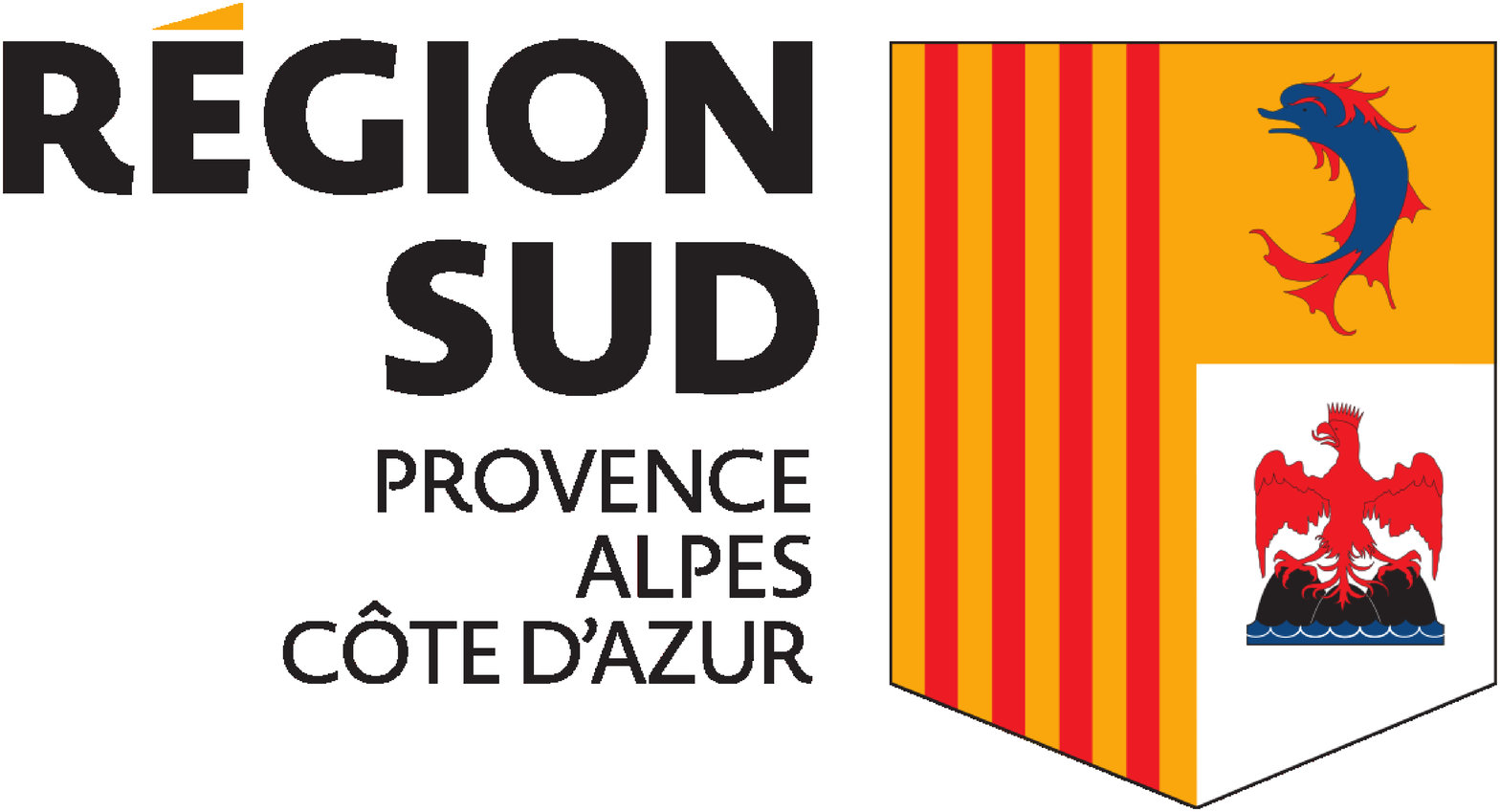}
\end{subfigure}
\end{figure}
\printbibliography
\end{document}